\documentclass[10pt]{IEEEtran}

\ifCLASSINFOpdf

\fi

\usepackage{tcolorbox}
\usepackage[subrefformat=parens,labelformat=parens,caption=false,font=footnotesize]{subfig}
\usepackage{array}
\usepackage{balance}
\usepackage{amsmath}
\usepackage{breqn}
\usepackage[nolist,nohyperlinks]{acronym}
\usepackage{graphicx,wrapfig,lipsum}
\usepackage{rotating}
\usepackage{amsmath, amssymb, amsthm}
\usepackage{stmaryrd}
\usepackage{tikz}
\usepackage{hyperref}
\usepackage{verbatim}
\usepackage{url}
\usepackage[utf8]{inputenc}
\usepackage[english]{babel}

\usepackage{booktabs}

\usepackage{multicol}
\usepackage{xr-hyper}

\usepackage{scalerel}

\usepackage[noadjust]{cite}
\usepackage[normalem]{ulem}
\usepackage{dsfont}
\usepackage{bm}
\usepackage[short]{optidef} 
\usepackage{diagbox}
\usepackage{enumitem}
\usepackage{slashbox}
\usepackage[ruled]{algorithm2e}
\usepackage{multirow}
\usepackage[T1]{}
\usepackage{color, colortbl}
\usepackage{babel}
\usepackage{verbatim}
\usepackage{textcomp}
\usepackage{tabulary}
\usepackage{booktabs}
\usepackage{caption}
\usepackage{listings}
\definecolor{Gray}{gray}{0.95}
\definecolor{LightCyan}{rgb}{0.8,0.85,1}
\definecolor{LightBlue}{rgb}{0.6,0.6,1}
\usepackage[font=small]{caption}
\usepackage{mathtools}

\definecolor{GPT35context}{HTML}{F1B814}
\definecolor{GPT4}{HTML}{BD1E51}
\definecolor{GPT35}{HTML}{80ADCC}
\definecolor{constructCluster}{HTML}{2B83BA}

\definecolor{niceblue}{HTML}{007ED6}

\definecolor{viridisStart}{rgb}{0.3137, 0.1412, 0.4549}    
\definecolor{viridisMiddle}{rgb}{0.2510, 0.2980, 0.5490}    
\definecolor{viridisEnd}{rgb}{0.1882, 0.4549, 0.5490}       
\definecolor{synExample}{rgb}{0.1569, 0.5804, 0.5490}  

\newtcolorbox[auto counter, number within=section]{mytheorem}[2][]{colframe=blue!50!black, colback=blue!10, coltitle=black, fonttitle=\bfseries, title=#2,#1}

\definecolor{MidTone}{RGB}{192, 30, 10}

\setlist{nosep}
\newcommand\blfootnote[1]{%
  \begingroup
  \renewcommand\thefootnote{}\footnote{#1}%
  \addtocounter{footnote}{-1}%
  \endgroup
}
\usepackage{mathtools}

\makeatletter
\def\blfootnote{\gdef\@thefnmark{}\@footnotetext}
\makeatother

\begin{document}

\title{TeleMath: A Benchmark for Large Language Models in Telecom Mathematical Problem Solving}

\author{
Vincenzo Colle$^{*\dagger}$, Mohamed Sana$^{\dagger}$, Nicola Piovesan$^{\dagger}$, Antonio De Domenico$^{\dagger}$, Fadhel Ayed$^{\dagger}$, Merouane Debbah$^\ddagger$\\
$^{\dagger}$Paris Research Center, Huawei Technologies, Boulogne-Billancourt, France\\
$^{*}$Università degli Studi di Cassino e del Lazio Meridionale, Cassino, Italy\\
$^\ddagger${Khalifa University of Science and Technology, Abu Dhabi, UAE}\\
}

\maketitle

\thispagestyle{empty}

\begin{abstract}
The increasing adoption of artificial intelligence in telecommunications has raised interest in the capability of Large Language Models (LLMs) to address domain-specific, mathematically intensive tasks. Although recent advancements have improved the performance of LLMs in general mathematical reasoning, their effectiveness within specialized domains, such as signal processing, network optimization, and performance analysis, remains largely unexplored.
To address this gap,
we introduce \emph{TeleMath}, the first benchmark dataset specifically designed to evaluate LLM performance in solving mathematical problems with numerical solutions in the telecommunications domain. Comprising 500 question-answer (QnA) pairs, TeleMath covers a wide spectrum of topics in the telecommunications field. 
This paper outlines the proposed QnAs generation pipeline, starting from a selected seed of problems crafted by Subject Matter Experts.
The evaluation of a wide range of open-source LLMs reveals that best performance on TeleMath is achieved by recent models explicitly designed for mathematical or logical reasoning. In contrast, general-purpose models, even those with a large number of parameters, often struggle with these challenges. We have released the dataset\footnote{{https://huggingface.co/datasets/netop/TeleMath}} and the evaluation code to ease result reproducibility and support future research.
\end{abstract}


\section{Introduction}
\label{sec:intro}
As the telecom industry advances toward next-generation networks, with 5G and the upcoming 6G, \ac{AI} and \ac{ML} are expected to play an increasingly significant role. Within this evolving landscape,
\acp{LLM} have emerged as powerful tools for assisting and automating complex tasks in diverse technical fields thanks to their semantic and reasoning abilities. Importantly, \acp{LLM} have improved significantly in areas such as arithmetic, algebra, and more broadly, mathematical reasoning, largely driven by the scale and diversity of their training data. Moreover, advancements in prompting techniques \cite{CoT} \cite{2022arXiv220510625Z} and reasoning strategies via Reinforcement Learning (RL) have further enhanced their capabilities, enabling them to tackle increasingly complex and abstract challenges \cite{DeepSeek-R1}. 
\noindent
In the telecom-domain, \acp{LLM} are explored for automatic generation of code, protocols \cite{Hermes} \cite{liu2025} 
and network configurations \cite{wangNeworkHumanFriendly} \cite{VPP}. Furthermore, researchers are investigating whether \acp{LLM} are also effective in sophisticated optimization \cite{LLM-as-Optim} and forecasting problems \cite{TrafficLLM}. 
To realize these tasks, \acp{LLM} must possess a deep understanding of the underlying mathematical principles that govern them.

Despite recent efforts in evaluating \acp{LLM} on broad-view mathematical problems -- see \texttt{MATH} \cite{MATH} and \texttt{GSM8K} \cite{GSM8K} -- and telecom-related tasks, such as protocol summarization \cite{SPEC5G},  standard document  classification \cite{TelecomGPT} and general telecom knowledge \cite{TeleQnA}, a comprehensive assessment of the \acp{LLM} mathematical capabilities within the telecom-domain, which often require not only numerical precision but also domain-specific knowledge, remains less understood.
Although a recent work has explored the \ac{LLM} abilities in problem modeling and equation completion for the telecom domain \cite{li2025wirelessmathbench}, the challenging skill of solving mathematical problems, has not received any attention yet. 

This paper addresses this specific need by introducing TeleMath, the first benchmark dataset designed to evaluate \ac{LLM} capabilities in solving mathematical problems in the telecommunications domain. The contributions of this paper are as follows:
\begin{enumerate}
\item We introduce a novel framework to generate synthetic \ac{QnA} pairs on different categories of mathematical problems. Starting from an initial {set} of mathematical problems -- a seed dataset -- generated by \acp{SME}, our framework first extracts reusable problem blueprints from each element of the seed dataset; then, it uses the blueprints to produce a large set of synthetic \acp{QnA}. This approach facilitates the scalability of the entire process and makes our framework suitable to design synthetic dataset for different categories and domains. 

\item Building upon the data augmentation framework, we curate and publicly release TeleMath, a dataset consisting of 500 \acp{QnA} to assess \ac{LLM} ability to solve mathematical problems in the telecoms domain.

\item Finally, we benchmark leading open-source \acp{LLM} on TeleMath. This evaluation shows that recent reasoning-oriented models, explicitly designed to think step by step and explore multiple approaches, perform significantly better than general-purpose models.
\end{enumerate}
\vspace{0.2cm}

\section{Dataset Sources And Characteristics}
\label{sec:sources}
This section introduces the seed dataset, which forms the backbone for building TeleMath. Next, we highlight the characteristics of TeleMath and we present the rationale behind our design choices.
\subsection{TeleMath Seed Dataset}
We generate and curate the seed dataset with the help of 10 \acp{SME}, who designed an initial set of 50 problems covering diverse areas, from fundamental concepts to more advanced topics in the telecommunications field. These problems span various difficulty levels to ensure that the dataset is comprehensive and reflects the variety of telecommunications domains. In addition, for each problem, the \acp{SME} provided, together with the solution, detailed step-by-step explanation, in symbolic or textual form, to enable effective generation of the TeleMath dataset.

\begin{figure}[!t]
    \centering
\includegraphics[width=0.48\textwidth]{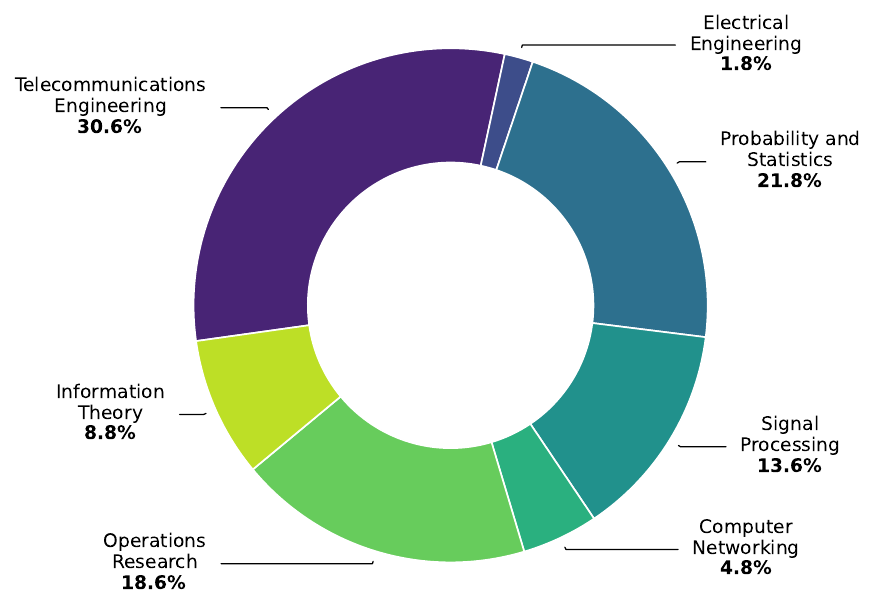}
    \caption{Distribution of TeleMath \acp{QnA} across their categories.}
    \label{fig:datasetdescription}
\end{figure}

\subsection{Dataset Characteristics}
\label{sec:characteristics}

\noindent
 TeleMath adopts a \ac{QnA}  format where answers are strictly numerical quantities.  Although telecommunications problems often involve symbolic expressions, we prioritize numerical answers for two reasons:  

\begin{itemize}  
    \item \textbf{Validation Robustness:} Numerical answers enable simple and clear evaluation. Answers containing formulas introduce variability e.g., due to the generation of expressions equivalent to the ground truth but different from a syntactical point of view, which may trigger false negatives during automated grading, even when employing Computer Algebra Systems such as Mathematica and SymPy.

    \item \textbf{Domain-Specific Practicality:} In telecommunications engineering, numerical results (e.g., signal-to-noise plus interference ratios) are often the primary input for real-world decision-making (e.g., antenna placement, network optimization, and protocol configuration). By focusing on numerical answers, the dataset mirrors the types of calculations engineers regularly perform, ensuring that models are trained to generate practical, actionable results.\\[0.1em]
\end{itemize}
\noindent
\textbf{Dataset Format.}
To maintain consistency across the dataset, we standardize the format for each \ac{QnA} in JSON, with the following fields:

\begin{itemize}
    \item \textbf{Question:} A question designed to test a specific concept within the telecommunications domain, framed so that the answer will always be a numerical value.
    
    \item \textbf{Answer:} The correct numerical answer to the question, represented either as a decimal or integer number, depending on the required quantity. The unit of the answer is either explicitly stated in the question or clearly implied by context, thus avoiding the need for unit conversion during validation, reducing the risk of errors.
    
    \item \textbf{Category:} A label classifying the \ac{QnA} within a specific domain of telecommunications. Figure~\ref{fig:datasetdescription} provides an overview of TeleMath category distribution. 
    
    \item \textbf{Tags:} Keywords that further categorize the \ac{QnA}, allowing for  granular classification. For example, a question about SNR might be tagged with ``SNR'', ``signal processing'', and ``noise''.
    
    \item \textbf{Difficulty:} The difficulty level of the question, classified as either Basic or Advanced. 
    
\end{itemize}

\section{TeleMath Dataset Creation}  
\label{sec:synthetic_data}

 The full generation of hundreds of problems by dedicated SMEs with step-by-step solutions and formula derivations would be prohibitively expensive and time-consuming. To address this challenge, we develop a framework composed of two synthetic data generation pipelines that expand a small seed dataset of problems generated by \acp{SME}. These two pipelines serve complementary roles: one is dedicated to generating blueprints for problems with numerical answers, while the other handles problems whose solutions are equations. Together, they enable the generation of large volumes of high-quality data in scalable manner.   

\begin{figure*}
    \includegraphics[width=1\textwidth]{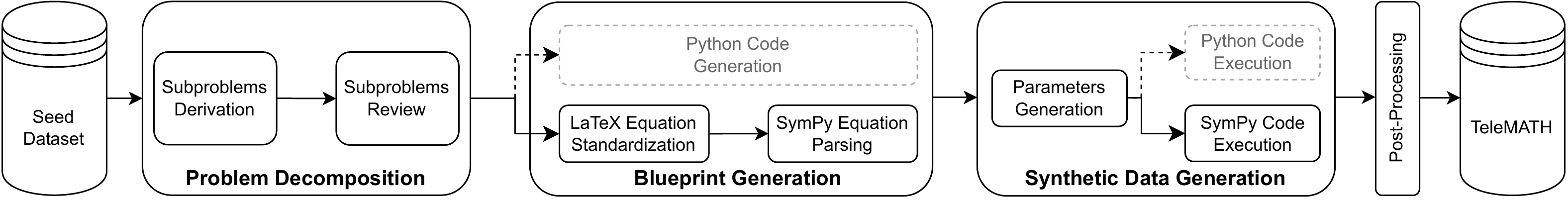}
    \caption{
    A high-level overview of the pipelines for Q\&A generation, with the symbolic math-driven path highlighted.}
    \label{fig:generationprocess}
\end{figure*}

 Figure \ref{fig:generationprocess} shows the overall dataset creation framework composed by four key blocks, namely: problem decomposition, blueprint generation, synthetic data generation, and post-processing. The following subsections elaborate on each of these blocks.

\subsection{Problem Decomposition}
\label{sec:prob_decomposition}
\label{subsec:subquestion_extraction}

In this block, we employ Qwen2.5-72B-Instruct, chosen for its strong instruction adherence capabilities, to break down complex \ac{SME} problems into more granular units, i.e., subproblems.
We prompt the LLM with the original problem, its solution, and the complete, step-by-step explanation. The problem decomposition is made of the two following phases: 
\begin{enumerate}
    \item \textbf{Subproblems Derivation:} In this task, the \ac{LLM} analyzes the problems provided by the SMEs and identifies individual steps, or short sequences of logically connected steps, to develop self-contained subproblems. The \acp{SME} manually numbered each step in the solutions of their problems, aiding the \ac{LLM} to fulfill this task. For each identified subproblem, the model is also instructed to derive the corresponding solution, inferring it from the original problem. 
    \item \textbf{{Subproblems} Review:} In this task the \ac{LLM} verifies that each derived subproblem is well-posed, unambiguous, and self-contained. Importantly, with this process, we confirm that each subproblem does not require additional information from the original SME problem.
\end{enumerate}

\subsection{Blueprint Generation}
The core of the question generation methodology lies in creating versatile ``blueprints'', either in the form of executable code or symbolic mathematical formulas, based on the derived subproblem. These blueprints can be instantiated to produce a diverse and large set of \acp{QnA}. 

\begin{figure}[!t]
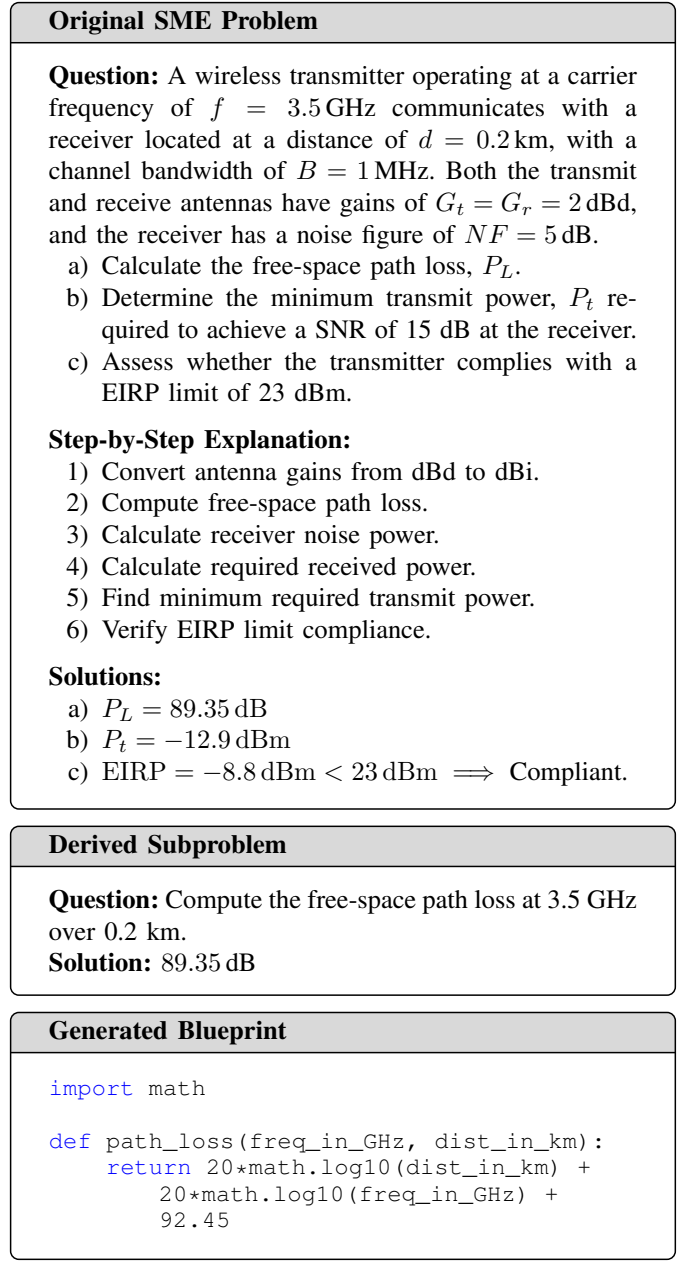

\begin{tcolorbox}[title=\textbf{Original SME Problem},
  colframe=black,
  colback=white,
  coltitle=black,
  colbacktitle=gray!30,
  fonttitle=\bfseries,
  boxrule=0.2mm,
  width=\linewidth, 
]
\textbf{Question:}  
A wireless transmitter operating at a carrier frequency of \( f = 3.5\,\text{GHz} \) communicates with a receiver located at a distance of \( d = 0.2\,\text{km} \), with a channel bandwidth of \( B = 1\,\text{MHz} \). Both the transmit and receive antennas have gains of \( G_t = G_r = 2\,\text{dBd} \), and the receiver has a noise figure of \( NF = 5\,\text{dB} \).
\begin{enumerate}[label=\alph*)]
    \item Calculate the free-space path loss, $P_L$.
    \item Determine the minimum transmit power, $P_t$ required to achieve a SNR of 15 dB at the receiver.
    \item Assess whether the transmitter complies with a EIRP limit of 23 dBm.
\end{enumerate}

\vspace{0.2cm}
\textbf{Step-by-Step Explanation:}
\begin{enumerate}
    \item Convert antenna gains from dBd to dBi.
    \item Compute free-space path loss. 
    \item Calculate receiver noise power.
    \item Calculate required received power.
    \item Find minimum required transmit power. 
    \item Verify EIRP limit compliance. 
\end{enumerate}
\vspace{0.2cm}
\textbf{Solutions:}
\begin{enumerate}[label=\alph*)]
\item $P_L = 89.35\,\mathrm{dB}$
\item $P_t = -12.9\,\mathrm{dBm}$
\item $ \mathrm{EIRP} = -8.8\,\mathrm{dBm} < 23\,\mathrm{dBm} \implies \text{Compliant.}$
\end{enumerate}
\end{tcolorbox}

\begin{tcolorbox}[title=\textbf{Derived Subproblem},     colframe=black,              
  colback=white,               
  coltitle=black,              
  colbacktitle=gray!30,        
  fonttitle=\bfseries,
  boxrule=0.2mm,               
  width=\linewidth, 
]
\textbf{Question:} Compute the free-space path loss at 3.5 GHz over 0.2 km.

\textbf{Solution:} \( 89.35\,\text{dB} \) 
\end{tcolorbox}

\begin{tcolorbox}[
  title=\textbf{Generated Blueprint},
  colframe=black,              
  colback=white,               
  colbacktitle=gray!30,        
  coltitle=black,              
  fonttitle=\bfseries,
  boxrule=0.2mm,               
  width=\linewidth,            
  top=0pt,
  bottom=0pt,
]
\begin{lstlisting}[language=Python, framextopmargin=0pt]
import math

def path_loss(freq_in_GHz, dist_in_km):
    return 20*math.log10(dist_in_km) + 20*math.log10(freq_in_GHz) + 92.45
\end{lstlisting}
\end{tcolorbox}

\caption{Example of an original problem, together with one of the derived subproblems and the associated blueprint.}

\label{fig:original_problem_and_subproblem}

\end{figure}

We define two approaches to derive blueprints from subproblems derived in the problem decomposition phase:
\begin{enumerate}
    \item \textbf{Code-Driven Blueprint Generation:}
    This approach converts subproblems for which the expected answer is a numerical value into executable Python codes. For this task, we select Qwen2.5-Coder-32B-Instruct, for its strong performance on code generation benchmarks. The model is provided with the subproblem along with the full step-by-step explanation, which serve as context to generate the corresponding Python code, i.e., the code-driven blueprint. The code is then executed with the original values of the subproblem parameters, and its output is compared with the expected result. If there is a mismatch, the generated code blueprint is discarded. Figure \ref{fig:original_problem_and_subproblem} shows an example of a generation of a Python blueprint, including the original problem and the derived subproblem. 
    
    \item \textbf{Symbolic Math-Driven Blueprint Generation:}
    For subproblems where the solution involves deriving an equation, we extract the underlying symbolic expression and transform it into a reusable blueprints for generating numerical \acp{QnA}. Given that symbolic expressions require rigorous structural consistency for reliable parsing, the process begins with standardizing the relevant portion of the solution from the original problem. This is handled by Qwen2.5-72B-Instruct, which reformats the mathematical expressions into a consistent LaTeX representation.
    
    The LaTeX code is then parsed using SymPy, a Python library for symbolic mathematics, which converts it into a structured algebraic form with free parameters, i.e., the symbolic blueprint.
\end{enumerate}

\subsection{Synthetic Data Generation}
From the generated blueprints, new QnAs can be created by defining new input parameters. To do so, for each subproblem, we instruct an LLM, Qwen2.5-72B-Instruct, to generate plausible inputs for the generated blueprints using the original problem and step-by-step solution as a context. This contextual information provides strong reference to the LLM for proposing realistic and well-defined input parameters.

For example, if a subproblem involves calculating signal loss for a 5G small cell at 3.5 GHz over a certain distance, the model, drawing on its vast training data, will propose other realistic frequencies and distances suitable for similar small cell scenarios, rather than arbitrary numbers. This ensures that the new parameters are contextually appropriate for the problem domain. Then, the blueprint is instantiated with the proposed parameters by executing the generated Python code for code-driven blueprints and the Sympy code for symbolic blueprints, respectively.

\begin{figure}
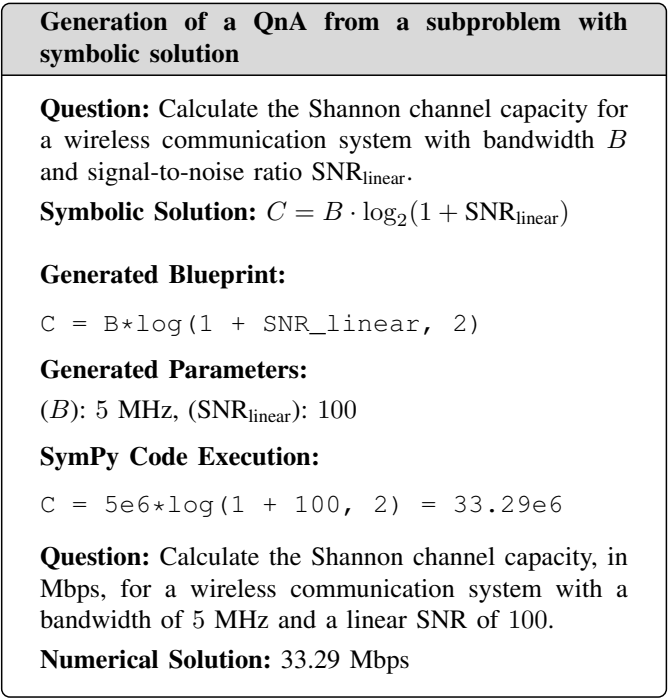

\centering
\begin{tcolorbox}[
  title=\textbf{Generation of a \ac{QnA} from a {subproblem} with symbolic solution}, 
  colframe=black,              
  colback=white,              
  coltitle=black,             
  colbacktitle=gray!30,       
  fonttitle=\bfseries,
  boxrule=0.2mm,              
  width=\linewidth
]
\textbf{Question:} Calculate the Shannon channel capacity for a wireless communication system with bandwidth $B$ and signal-to-noise ratio $\text{SNR}_{\text{linear}}$. \\[0.3em]
\textbf{Symbolic Solution:} $C = B \cdot \log_2(1 + \text{SNR}_{\text{linear}})$ \\

\textbf{Generated Blueprint:}
\begin{verbatim}
C = B*log(1 + SNR_linear, 2)
\end{verbatim}

\textbf{Generated Parameters:} \\[0.3em]
($B$): $5 \text{ MHz}$, ($\text{SNR}_{\text{linear}}$): $100$

\vspace{0.2cm}
\textbf{SymPy Code Execution:}
\begin{verbatim}
C = 5e6*log(1 + 100, 2) = 33.29e6
\end{verbatim}
\vspace{0.2em}
\textbf{Question:} Calculate the Shannon channel capacity, in Mbps, for a wireless communication system with a bandwidth of $5 \text{ MHz}$ and a linear SNR of $100$. \\[0.3em]
\textbf{Numerical Solution:} 33.29 \text{Mbps}
\end{tcolorbox}
\caption{Generation of a \ac{QnA} from a {subproblem} with symbolic solution.}
\label{fig:rate-transformation}
\end{figure}

\subsection{Post-Processing}

Following the data generation step, the process transitions into the post-processing stage. This phase ensures that the generated data is both usable and consistent with the original problem structure, which is realized by three functions:

\begin{enumerate}
\item \textbf{Filtering:}
This step eliminates any generated outputs that fall outside acceptable bounds. For example, excessively large or small values for distance, frequency, or power are often indicative of invalid input parameters. By enforcing domain-specific thresholds, it is ensured that only realistic and physically plausible problem instances are retained.

\item \textbf{Question Editing:}  
The numerical parameters defined in the data generation step are injected into the original question. This editing is performed by prompting {Qwen2.5-72B-Instruct} with both the original question and a dictionary of the generated parameter names and their corresponding values. The model is instructed to substitute the symbols and the values in the original question with the provided values, preserving the original structure of the text. 
Figure \ref{fig:rate-transformation} shows an example of the generation of a new question from a subproblem with symbolic solution.

\item \textbf{Semantic Validation:}  
A dedicated validator powered by {Qwen2.5-72B-Instruct} is employed to compare each rewritten question against its original counterpart. The goal is to ensure semantic fidelity by verifying:

\begin{itemize}[left=0pt]
    \item \textbf{Structural Equivalence:} The underlying problem structure and required solution steps remain unchanged.
    \item \textbf{Dimensional Consistency:} All physical units are preserved and logically consistent.
\end{itemize}
\end{enumerate}

After processing, each \ac{QnA} is annotated with auxiliary metadata -- category, tags, and difficulty -- assigned heuristically by Qwen2.5-Math-7B-Instruct. 
To do so, we prompt the model with the problem statement, instructing it to select a category and to propose tags that capture finer-grained concepts. The output categories are further clustered and each cluster is renamed to avoid an excessive number of categories in TeleMath.
The \ac{QnA} difficulty is determined by the abulity of Qwen2.5-Math-7B-Instruct to correctly answer it. The average token count of these solutions is used as a proxy for complexity. The distribution of the average token counts shows a bimodal pattern, naturally separating TeleMath problems into two groups: those with shorter solutions (fewer tokens, indicating lower complexity) are labeled as Basic, and those with longer solutions (more tokens, indicating higher complexity) are labeled as Advanced.
    
\begin{table*}[t!]
\centering
\resizebox{\textwidth}{!}{
\begin{tabular}{lccccccccc} 
\toprule
\diagbox{\textbf{Category}}{\textbf{Model}} 
  & \multicolumn{5}{c}{\textbf{Reasoning}}
  & \multicolumn{4}{c}{\textbf{Non-reasoning}} \\
\cmidrule(lr){2-6} \cmidrule(lr){7-10}
  &
  \shortstack{\textbf{Metric}} 
  & 
  \shortstack{\textbf{Qwen3}\\\textbf{32B}} 
  & 
  \shortstack{\textbf{DeepSeek-R1}\\\textbf{Distill-Llama-70B}} 
  & 
  \shortstack{\textbf{Phi-4}\\\textbf{Reasoning+}} 
  & 
  \shortstack{\textbf{Qwen3}\\\textbf{4B}} 
  & 
  \shortstack{\textbf{Qwen2.5}\\\textbf{Math-72B*}} 
  & 
  \shortstack{\textbf{Llama-3.3}\\\textbf{70B*}} 
  & 
  \shortstack{\textbf{Qwen2.5}\\\textbf{Math-7B*}} 
  & 
  \shortstack{\textbf{Llama-3.1}\\\textbf{8B*}} \\ 
\midrule

\multirow{2}{*}{Computer Networking [CN]}
 & \texttt{pass@1} 
 & 55.99 & 47.66 & 30.99 & 14.32 & 26.61 & 26.30 & 6.51 & 4.95 \\
 & \texttt{cons@16}
 & 66.67 & 54.17 & 29.17 & 12.50 & 32.26 & 29.17 & 12.50 & 0.00 \\
\addlinespace[8pt]

\multirow{2}{*}{Electrical Engineering [EE]} 
 & \texttt{pass@1} & 72.92 & \textbf{72.92} & 66.67 & \textbf{65.28} & \textbf{55.80} & \textbf{63.19} & 27.78 & \textbf{34.03} \\
 & \texttt{cons@16} & 77.78 & \textbf{77.78} & 77.78 & 66.67 & \textbf{64.29} & \textbf{66.67} & 33.33 & \textbf{55.56} \\
\addlinespace[8pt]

\multirow{2}{*}{Information Theory [IT]} 
 & \texttt{pass@1} & 76.99 & 62.07 & \textbf{70.74} & 64.77 & 39.70 & 38.21 & 27.98 & 13.64 \\
 & \texttt{cons@16} & \textbf{81.82} & 75.00 & \textbf{79.55} & \textbf{72.73} & 46.48 & 36.36 & 31.82 & 22.73 \\
\addlinespace[8pt]

\multirow{2}{*}{Operations Research [OS]} 
 & \texttt{pass@1} & 70.63 & 55.98 & 54.17 & 49.40 & 52.39 & 40.99 & 27.22 & 14.45 \\
 & \texttt{cons@16} & 72.04 & 64.52 & 56.99 & 53.76 & 58.26 & 50.54 & 26.88 & 22.58 \\
\addlinespace[8pt]

\multirow{2}{*}{Probability \& Statistics [PS]} 
 & \texttt{pass@1} & \textbf{77.47} & 70.81 & 67.60 & 59.58 & 49.49 & 49.77 & \textbf{34.52} & 16.40 \\
 & \texttt{cons@16} & 80.73 & 75.23 & 71.56 & 63.30 & 52.59 & 56.88 & \textbf{39.45} & 22.02 \\
\addlinespace[8pt]

\multirow{2}{*}{Signal Processing [SP]} 
 & \texttt{pass@1} & 71.05 & 43.93 & 52.39 & 41.18 & 36.11 & 32.17 & 21.14 & 15.63 \\
 & \texttt{cons@16} & 77.94 & 55.88 & 63.24 & 50.00 & 45.56 & 39.71 & 27.94 & 25.00 \\
\addlinespace[8pt]

\multirow{2}{*}{Telecom. Engineering [TE]} 
 & \texttt{pass@1} & 62.25 & 40.28 & 41.54 & 33.62 & 30.50 & 24.88 & 11.89 & 10.21 \\
 & \texttt{cons@16} & 73.86 & 46.41 & 45.10 & 36.60 & 38.05 & 23.53 & 16.99 & 15.03 \\
\midrule
\rowcolor{gray!10}
\multicolumn{10}{l}{\textbf{Overall Performance}} \\
\midrule
\multirow{2}{*}{Accuracy}
    & \texttt{pass@1}   & 69.51  & 53.21
    & 53.56
    & 45.62
    & 39.99
    & 36.23
    & 22.38
    & 13.56\\
    & \texttt{cons@16}  & 
    76.00 & 
    60.80 & 58.40 & 50.00 & 46.48 & 
    40.20 & 26.60 & 20.00 \\
\addlinespace[8pt]
\multirow{1}{*}{Top Domain} 
    & \texttt{pass@1}   & PS      & EE      & IT      & EE      & EE      & EE      & PS      & EE      \\
\bottomrule
\end{tabular}
}
\caption{Performance comparison of pass@1 and cons@16 accuracy. Asterisks (*) denote Instruct variants (e.g., Qwen2.5-72B-Math* stands for Qwen2.5-72B-Math-Instruct).}
\label{table:benchmarks}
\end{table*}

\section{Performance Evaluation}

\noindent In this section, we present the performance of popular open-source language models on TeleMath.
In our experiments, we generate $N=16$ independent responses for each question and model pair. This generation process uses a sampling temperature of 0.6, a top-$p$ value of 0.90 and a maximum generation length of 16,384 tokens for all models, except for Qwen2.5-Math-72B-Instruct and Qwen2.5-Math-7B-Instruct, which are limited to a maximum generation length of 4,096 tokens. 

From the generated responses, we evaluate the model performance on TeleMath using two metrics: \texttt{pass@1} (pass at 1) and \texttt{cons@16} (consensus at 16):

\noindent
\textbf{\texttt{pass@1}:} This metric assesses the ability of the model to generate a correct answer within a single attempt, averaged over multiple sample answers.

\noindent
\textbf{\texttt{cons@16}:} This metric evaluates performance based on majority voting over the 16 generated answers. The problem is considered correctly solved if the most frequently occurring answer among the generated ones is correct.

\subsection{Models Performance Analysis}
\label{subsec:perfbenchmark}
Table \ref{table:benchmarks} summarizes the performance of various reasoning and non-reasoning models on TeleMath. Specifically, this table presents the pass@1 and cons@16 accuracies achieved by each model across the different \ac{QnA} categories of TeleMath. 

\noindent
{\textbf{Reasoning Models.}} Our benchmarking shows that Qwen3-32B outperforms all other models achieving an average pass@1 of 69.51\% and cons@16 of 76\%.
Following Qwen3-32B, DeepSeek-R1-Distill-Llama-70B (pass@1 53.21\%, cons@16 60.80\%) and Phi-4-reasoning+ 
(pass@1 53.56\%, cons@16 58.40\%) form a distinct second tier. It is worth to notice the performance of the smallest reasoning model, Qwen3-4B
, which surpasses that of substantially larger non-reasoning models. This finding underscores the impact of architecture and training methodologies optimized for reasoning tasks. Such optimization allows even more compact models to outperform larger, general-purpose counterparts in these specialized evaluations. For instance, in Information Theory, Qwen3-4B clearly outperforms Llama-3.3-70B-Instruct and Qwen2.5-Math-72B-Instruct, despite being approximately 17 times smaller.\\[0.2em]
\noindent
{\textbf{Non-Reasoning Models.} On the other hand,} “Non-reasoning” models generally lag behind ``reasoning models". Within the former group, Qwen2.5-Math-72B-Instruct leads with an overall pass@1 of 39.99\% and cons@16 of 46.48\%, followed by Llama-3.3-70B-Instruct at 36.23\% (cons@16 40.20\%). It is noteworthy that the “Math” specialized models (Qwen2.5-Math-72B-Instruct and Qwen2.5-Math-7B-Instruct) operate with a constrained maximum completion length of 4096 tokens. Despite this limitation, Qwen2.5-Math-72B-Instruct emerges as the best non-reasoning model. This suggests its specialized mathematical training can yield acceptable results even within such constraints. However, small non-reasoning models like Qwen2.5-Math-7B-Instruct (pass@1 22.38\%, cons@16 26.60\%) and Llama-3.1-8B-Instruct (pass@1 13.56\%, cons@16 20\%) exhibit considerably lower performance across the board. \\[0.2em]
\noindent
\textbf{Model Scale.} The influence of model scale is consistently evident within model families. For instance, Qwen3-32B significantly outperforms Qwen3-4B. Similar trends are observed for the Qwen2.5-Math (72B vs. 7B) and Llama-3.x (70B vs. 8B) series. This observation reinforces the general principle that larger models within the same architectural family tend to yield better results. \\[0.2em]
\noindent
\textbf{Domain Performance.}
Domain-wise, Electrical Engineering emerges as a field where several models achieve strong scores. This includes the non-reasoning Llama-3.3-70B-Instruct (pass@1 63.19\%, cons@16 66.67\%), which is more than 25 percentage points higher than the overall average. Such performance is likely due to greater representation of Electrical Engineering content in the training data, compared to more specialized fields. In contrast, Computer Networking and Telecommunications Engineering proved to be more challenging for most models. This is especially true for smaller non-reasoning models, where scores often fall below 20\%. \\[0.2em]
\noindent
\noindent
In summary, the analysis highlights the current superiority of reasoning-focused architectures. 
Among the evaluated models, Qwen3-32B consistently achieves the highest average performance in each category. The results also suggests that for complex technical domains requiring nuanced understanding and problem-solving, models designed and trained for reasoning offer a clear advantage. This holds true even when comparing smaller reasoning models to larger non-reasoning counterparts. Although large general-purpose instruct models and specialized mathematical models show promise and utility, they generally do not reach the same level of consistent high performance as the top-tier reasoning models.

\section{Limitations}

TeleMath represents a significant step forward in evaluating LLMs on telecommunications-specific mathematical problems. 
However, despite our efforts to ensure broad coverage, the seed dataset (50 problems) does not fully represent the entire telecommunications domain. Moreover, the current distribution of problem categories (as shown in Fig.~1) skews toward traditional telecommunications engineering (30.6\%). Although this provides depth in a core area, it may also imply that other important categories within the telecommunications landscape are under-represented.

{Additionally, using Qwen family models  for SME problem decomposition, blueprint generation, and review may introduce some bias in the TeleMath dataset, favoring higher scores for these models. However, we have limited this bias through the semantic validation step, which aim at ensuring semantic fidelity of the generated QnAs with respect to the original problems.}

\section{Conclusions}
\label{sec:conclusions}
This paper introduces TeleMath, a novel benchmark dataset specifically developed to assess the mathematical problem-solving abilities of \acp{LLM} within the telecommunications domain. Recognizing a gap in evaluating LLMs on specialized, quantitative tasks, TeleMath offers 500 curated question-answer pairs sourcing from expert knowledge augmented through robust synthetic generation pipelines. Our evaluation of diverse open-source LLMs on this dataset reveals a key insight: models explicitly designed for mathematical or logical reasoning consistently achieve superior performance compared to even larger, general-purpose counterparts. By making TeleMath publicly available, we provide a resource for the research community to rigorously benchmark, understand, and ultimately enhance LLM capabilities for the complex mathematical demands inherent in the telecommunications sector and similar technical fields.

\bibliographystyle{IEEEtran}

\bibliography{reference.bib}

\begin{acronym}[AAAAAAAAA]
  \acro{3GPP}{Third Generation Partnership Project}
  \acro{QnA}{Question and Answer}
 \acro{AI} {Artificial Intelligence}
 \acro{BERT}{Bidirectional Encoder Representations from Transformers}
  \acro{DL}{Deep Learning}
   \acro{FPGA}{Field-Programmable Gate Array}
  \acro{GPT}{Generative Pre-trained Transformer}
  \acro{KPI}{Key Performance Indicator}
 \acro{LLM}{Large Language Model}
  \acro{MNO}{Mobile Network Operator}
\acro{BS}{Base Station}
  \acro{ML}{Machine Learning}
 \acro{NLP}{Natural Language Processing} 
 \acro{API}{Application Programming Interface} 
 \acro{RAN}{Radio Access Network}
 \acro{RAG}{Retrieval Augmented Generation}
 \acro{MIMO}{multiple-input multiple-output}
 \acro{SME}{Subject Matter Expert}
 \acro{QnA}{question-answer}

 \end{acronym}

\end{document}